\documentclass[letterpaper]{article} 
\usepackage{aaai23}  
\usepackage{times}  
\usepackage{helvet}  
\usepackage{courier}  
\usepackage[hyphens]{url}  
\usepackage{graphicx} 
\usepackage{natbib}  
\usepackage{caption} 
\usepackage{algorithm}
\usepackage{algorithmic}

\usepackage{newfloat}
\usepackage{listings}
\DeclareCaptionStyle{ruled}{labelfont=normalfont,labelsep=colon,strut=off} 
\floatstyle{ruled}
\newfloat{listing}{tb}{lst}{}
\floatname{listing}{Listing}
\title{The Role of Heuristics and Biases During Complex Choices with an AI Teammate}
\author {
    Nikolos Gurney\textsuperscript{\rm 1},
    John H. Miller \textsuperscript{\rm 2,3},
    David V. Pynadath, \textsuperscript{\rm 1}
}
\affiliations {
    \textsuperscript{\rm 1} Institute for Creative Technologies, University of Southern California\\
    \textsuperscript{\rm 2} Carnegie Mellon University\\
    \textsuperscript{\rm 3} Santa Fe Institute\\
    gurney@ict.usc.edu, pynadath@ict.usc.edu, JM7T@andrew.cmu.edu
}

\usepackage{bibentry}

\begin{document}

\maketitle

\begin{abstract}
Behavioral scientists have classically documented aversion to algorithmic decision aids, from simple linear models to AI. Sentiment, however, is changing and possibly accelerating AI helper usage. AI assistance is, arguably, most valuable when humans must make complex choices. We argue that classic experimental methods used to study heuristics and biases are insufficient for studying complex choices made with AI helpers. We adapted an experimental paradigm designed for studying complex choices in such contexts. We show that framing and anchoring effects impact how people work with an AI helper and are predictive of choice outcomes. The evidence suggests that some participants, particularly those in a loss frame, put too much faith in the AI helper and experienced worse choice outcomes by doing so. The paradigm also generates computational modeling-friendly data allowing future studies of human-AI decision making.
\end{abstract}

\section{Introduction}

The superiority of passive, algorithmic decision making is dogma in the behavioral sciences (\citealt{dawes1974linear,dawes1989clinical}). This is not a surprise given that even poorly-tuned models outperform humans (\citealt{dawes1979robust}). Nevertheless, researchers have documented a strong aversion to relying on algorithms (\citealt{dietvorst2015algorithm, burton2020systematic}). There is increasing evidence, however, that people are more open to input from algorithmic decision aids (\citealt{logg2019algorithm}). This shift towards a more trusting stance aligns with intuition: combining the proliferation of algorithmic decision aids and the insight that repeated exposure to stimuli can alter the ways in which people respond to them (\citealt{zajonc1968attitudinal, bornstein1992stimulus}) creates a scenario in which it would be surprising if people \textit{did not} exhibit more trust. 

Increasingly, decision aids rely on adaptive rather than passive algorithms. These artificial intelligence (AI) helpers are optimized to function in settings where passive algorithms, such as linear models, underperform relative to human decision making. Although these technologies function and behave in fundamentally different ways from more passive predecessors, researchers largely still study them by relying on traditional empirical approaches: either simple choice experiments (e.g. \citealt{logg2019algorithm}, in which the authors use their evidence to motivate the need for further research, particularly in settings where AI are in use) or \textit{in situ} (in which there are entire sub-fields, such as human-robot interaction). We argue that a middle ground is needed where the experimental setting matches the technological capabilities of AI but avoids the contextual challenges of \textit{in situ} experimentation. 

We present results from just such an experiment in which participants work with an AI helper to search through complex choice options. We find classic behavioral biases, specifically anchoring and framing effects, are still present in the choice behavior. The biases are predictive of choice outcomes and highlight opportunities to develop AI helpers that can improve outcomes by accounting for these biases. Critically, we engineered the experimental environment such that it generates AI-friendly data. This feature creates future opportunities for training AI helpers using data from an experimental paradigm in which known artifacts of human cognition are present. 

\section{Related Works}
\subsection{Complex Choices}
A blossoming corner of behavioral science is working to understand the nuance of human judgment and decision making during \textit{complex choice}. A complex choice is one in which the relationships between the variables under consideration are nonlinear, which means that even when a choice is characterized by a small number of variables, their interactions can render optimization intractable \cite{gurney2022Complex}. Obvious examples of complex choices are home buying, career moves, and mate selection. In any of these examples, changing a single choice variable can dramatically impact how other variables affect the choice. In the case of buying a home, for example, the opportunity to work remotely changes how commuting distance, home size, and a host of other variables interact. It is not just monumental choices that suffer from complexity. Filling a shopping cart given a budget constraint, something that most adults regularly do, can be complex. A shopper must, for example, consider how the myriad of items they buy combine into meals, when they will perish, whether they will satisfy their taste preferences in the future, etc. 

Behavioral scientists have not ignored the unique nature of complex choices. Herbert Simon, for example, noted that many choices are simply intractable due to the interaction of variables (\citeyear{simon1956rational}). Despite this recognition, complex choice experiments have only recently emerged. Behavioral science historically relied on experimental designs that remove complexity from choices because it allows researchers to easily isolate behavioral nuances. These experiments usually take one of two forms: descriptive, in which all the information needed to make an informed choice is available, or experiential, in which participants must learn about the choice through repeated experience \cite{jessup2022choice}. 

In descriptive experiments, researchers typically manipulate a single choice feature in a between-subjects design. A classic example from studies of the anchoring heuristic is to ask people about the population density of a city after asking them if the population is more or less than a reference value \cite{jacowitz1995measures}. When the reference is low (high), people report that the city is more sparsely (densely) populated. Such experiments are crucial to dissecting human behavior, however, they are hamstrung by their rigidity. For example, it is well documented that human judgment and decision making is highly dependent on experience, including a person's own thoughts about such experiences \cite{schwarz2004metacognitive}. The very nature of complex choices renders descriptive experimentation, however, near impossible. Moreover, there is a growing body of work that suggests the instances in which it has been adapted may be undermined by other artifacts of human cognition (such as preference reversals) once people gain more experience \cite{hertwig2004decisions, tsetsos2012salience, erev2017anomalies, jessup2022choice}. 

The intractability of complex choice has contributed to the development of experiential study paradigms that adapt complex systems models. Two popular complex systems models are the ``multi-arm bandit'' \cite{berry1985bandit} and the $NK$ \cite{levinthal1997adaptation} paradigms. One experimental paradigm, for example, asks participants to combine $n$ shapes that interact $k$ ways to create art for aliens. This paradigm enabled the researchers to investigate how people search, explore, and work together during complex choices  \cite{billinger2014search, wu2018generalization, billinger2021exploration, billinger2022learning}. Although these experiments did not directly control for the bias, Billinger et al. (\citeyear{billinger2014search} did demonstrate that anchoring on past outcomes, using prior experience as an endogenous variable, can inform future complex choices). Study participants exerted more (less) effort when underperforming (overperforming) relative to their previous performance. 

\subsection{Complex Choice in Teams}
Teaming adds another element of complexity to any choice. In the alien art task, human dyads coordinated their search efforts during a complex choice without external guidance or resources through a mix of turn-taking and simultaneous moves \cite{billinger2022learning}. Teams can also facilitate better complex choices by expanding the sampled regions of a problem space. Each team member brings unique knowledge, preferences, and beliefs to the task, which their behavior may reflect. To illustrate, consider an experiment that tasked teams of participants with a simulated entrepreneurship task: managing a lemonade stand \cite{sommer2020you}. The teams worked together to learn about and combine the interacting variables, such as different ingredients, price levels, stand locations, etc., to make the lemonade stand profitable. When the task complexity was low, i.e. few interactions and relatively low stochasticity, interactive teams did better than teams of individual performers. This result flipped, however, as complexity increased. 

Behavioral scientists are not the only researchers working towards a deeper understanding of teaming during complex choices. The promise of human-machine teams capable of superior performance is one chased by researchers across computer science disciplines. The earliest examples from the literature are likely better described as cooperative efforts---such as airplane piloting and air-traffic control \cite{hoc2000human}. Early success in these and related domains birthed a sizable, ever-evolving sub-discipline that studies the dynamics of trust in automated systems \cite{lee2004trust}. This research area now spans settings from games that offer insights into the belief states of human teammates \cite{siu2021evaluation, chong2022human} to military scenarios that challenge common models of trust dynamics \cite{gurney2022measuring}. The expansive literature on human-AI teams now encompasses topics as diverse as what expectations people hold of AI teammates \cite{zhang2021ideal} to endowing AI with the ability to represent the belief states of their human counterparts \cite{gurney2022AI}. 

Conspicuously missing, however, from research on human-AI teaming is an investigation of what are arguably the most salient features of human judgment and decision making: heuristics and biases \cite{gigerenzer1999simple,kahneman2011thinking}. The obvious argument is that for an AI teammate to accurately model its human counterpart it needs to take into consideration the cognitive tools used by the human during choice. Some prior work on complex choice has endogenously tested the role of heuristics and biases, such as Billinger et al. (\citeyear{billinger2014search}) who demonstrated that anchoring on previous outcomes informs future search behavior. Building on this and other related research, our work exogenously controls for anchoring and framing effects, which are two of the most studied phenomena in the heuristics and biases literature. 

\subsection{Anchoring and Framing Effects}
Anchoring in human cognition is the tendency of final choices to reflect starting information, even if that information is uninformative \cite{tversky1974judgment, chapman1999anchoring, epley2006anchoring}. To illustrate, imagine walking down a store aisle and seeing a sign above a sale item proclaiming a given limit per customer. The anchoring effect suggests that you will buy more or less of the sale item depending on the stated limit despite your intentions or preferences \cite{wansink1998anchoring}. Similar effects are documented in a host of domains as diverse as retirement savings \cite{madrian2001power}, courtroom sentencing \cite{englich2006playing}, and even in mate selection \cite{goller2018anchoring}. 

Framing effects are similarly robust. The basic idea of framing is that for any prospect, a person has some reference value. Changes are evaluated relative to that value and differently if they are viewed as losses versus gains \cite{kahneman2011thinking}. The classic empirical demonstration of framing asks participants to consider two interventions to stop the spread of a deadly disease. All participants receive a probability-based option and an alternative. Half receive an alternative framed as a gain (lives saved), while the others receive one framed as a loss (lives lost) but that maintains the same ratio. The observation is a reversal in the number of participants selecting the probability-based option when the choice is framed as a loss \cite{tversky1985framing}. 

\section{Experimental Paradigm}
We adapted a rugged-landscape search metaphor, that was first introduced in biology to describe evolutionary selection \cite{wright1932roles}, in an interactive, complex choice task \cite{gurney2022method}. The fundamental idea of this metaphor is that evolutionary pressure, over many generations, guides organisms in a search for an optimal genotype across a genetic landscape where elevation on the landscape represents the fitness of a genotype. This metaphor was generalized in the $NK$ model \cite{kauffman1987towards, kauffman1989nk} and researchers have applied it to an array of different settings, perhaps most famously in the management literature \cite{levinthal1997adaptation}. Our task asks study participants to search such a landscape for its global peak (optimum) by tuning dials, each of which accesses a different landscape dimension. The complex choice that participants face is when to stop searching, i.e., what dial setting to submit as the best for a given landscape. Participants received a bonus payment for making better choices, i.e. submitting higher elevations. Each participant completed four dial tuning tasks, the first two on their own and then two more with an AI helper. Importantly, this task lacks strong contextual cues, unlike prior tasks (making artwork for aliens, managing a lemonade stand, farming, etc.).

\subsection{Landscapes as Choice Sets}
Landscapes are procedurally generated and unique to each participant. They vary in ruggedness, just like mountain ranges, so even a three-dimension task can be challenging for a person to solve. The only way that a person can know the value of a given location in the landscape is by visiting it. Depending on the smoothness of a landscape, a person may, however, make inferences about different regions based on information they previously uncovered. Smoothness is determined by the range of possible slopes between neighboring points in the landscape. For the present experiment, participants completed tasks with one (simple) or four (rugged) peaks and a slope setting such that two neighboring points never differed in value by more than 10\% of the absolute range of values, which was 33 units. The order of the tasks was randomized, but participants always did a simple and rugged task alone and with the AI helper.

Landscape topologies are created by randomly placing the global peak in a plane, which for this experiment encompassed 24 $\times$ 24 units. For multi-peaked landscapes, additional peaks are randomly placed by generating a list of candidate locations where they maintain a given level of prominence such that valleys separate peaks. The size and dimensionality of the landscape constrain the number of peaks and their elevations. Once all the peaks are placed, a smoothing algorithm ensures that the slopes between the peaks and valleys meet the needed constraints. Our experimental design implemented an elevation boundary of $[0, 32]$ and peaks ranging in the $[26, 32]$ interval with the tallest fixed at 32.

\begin{figure*}[ht]
\includegraphics[width=0.95\textwidth]{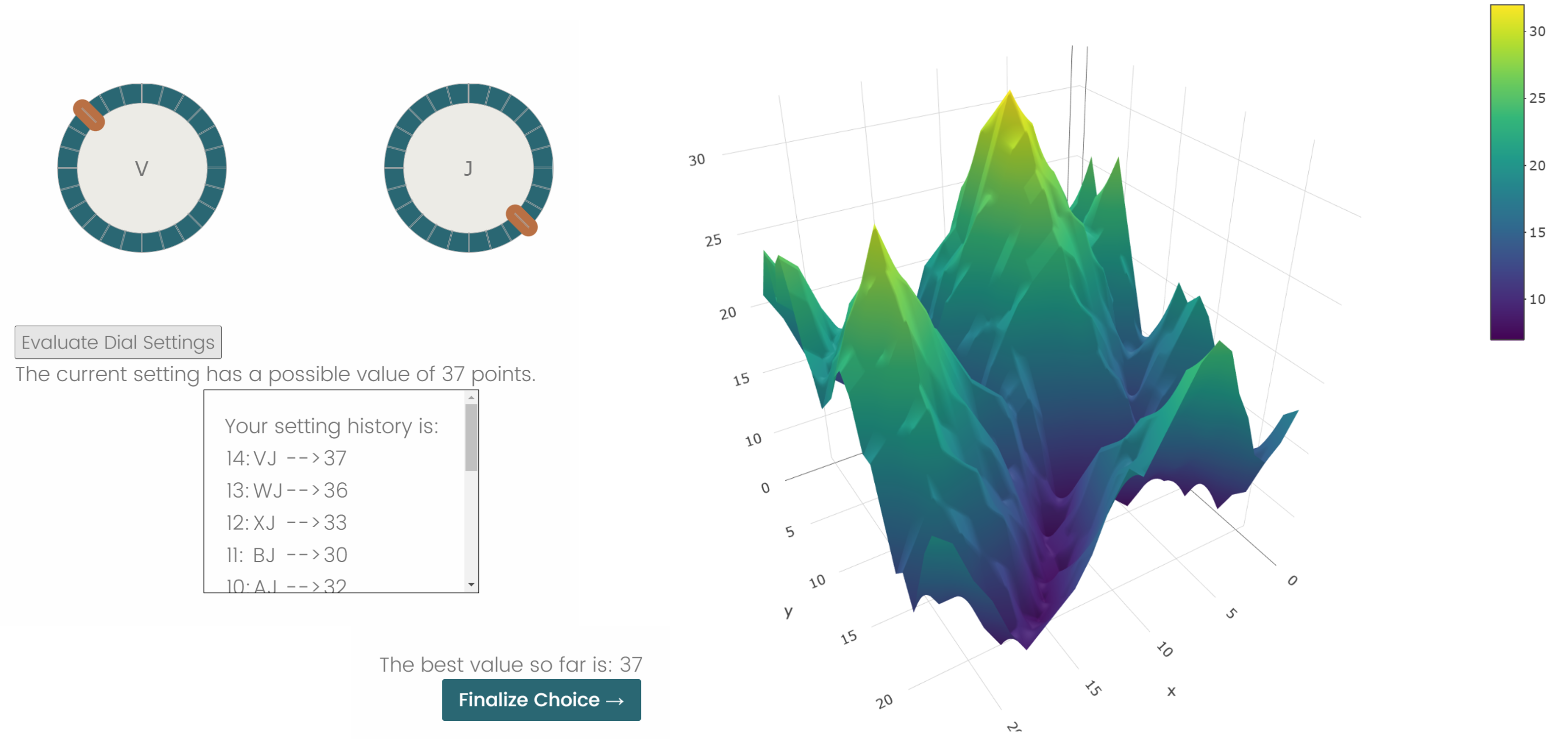}
\caption{A screenshot from the experiment is on the left, and a landscape rendering is on the right. All participants saw the same dial and feedback interface, but participants in the anchoring treatment saw the best value possible directly above the dials. A two-by-two image of the treatment conditions is available in the online supplement. Participants did not see a rendering of the landscape.}
\label{fig:screenshot}
\end{figure*}

The landscapes are rendered so that they are continuous over the edges, as in the right panel of figure \ref{fig:screenshot}. This means that rolling the east and west edges of a landscape together forms a smooth transition. The same is true of the north and south edges, thus each landscape forms a toroidal world characterized by peaks and valleys. Shifting that landscapes off of the $[0, 32]$ interval facilitates studying the impact of earlier trials on future behavior. The incentivized task is to find, via exploration, the tallest peak in a given world. In the right panel of figure \ref{fig:screenshot}, the global peak is the light yellow point in the upper center of the image. A dark blue valley is below it and surrounded by other peaks.

\subsection{User Interface}
Participants cannot see the landscapes. Instead, they uncover information about a landscape by ``tuning'' on-screen dials, as shown in the left panel of figure \ref{fig:screenshot}. Each dial is associated with a different dimension of the task. Participants submit tuned settings by clicking the ``Evaluate Dial Settings'' button. A two-dial experiment, for example, allows a person to explore a three-dimensional world, which is what we implemented for the present study. In this case, conceptually, one dial moves along the east-west and the other moves along the north-south dimension of the landscape. The feedback given when a setting is submitted is the elevation, or choice quality, of that dial combination. Importantly, the landscape dimensions, thus the dials, are interdependent and their relationships are nonlinear. 

Tuning happens by clicking on a dial handle and dragging it around the underlying ring. As the dial is moved, the letter in the center of the dial changes. Each position on a dial is associated with a letter, which we implemented to simplify feedback. Feedback about the historical dial settings is available in a numbered list below the dials. Participants are able to scroll through their history, see what settings they previously checked, and decide when they are ready to make their choice by clicking the ``Finalize Choice'' button. 

We implemented dials with 24 discrete settings. This matches the coordinate plane used to generate the landscapes. Our decision to use a plane of this size with fixed steps was based on pretesting that suggested such a landscape would yield a search space (576 unique combinations) that participants are unwilling to exhaustively search given the incentive structure. In other words, for most participants, uncertainty would remain around their choice.

We define two broad types of dial usage: exploring and exploiting. Exploration entails looking at regions of the landscape more than two units from any previously observed location. For example, if a participant observed the dial setting [A,A] on their first move, submitting [A,D] or [X,C] on the next move would constitute an explore. Exploitation is when a participant observes a new location less than three units from any previously observed setting. 

\subsection{AI Helper}
After completing two tasks, an AI helper took control of the right dial. It, like study participants, did not have direct access to the landscape, thus it learned about the topology through sampling. Importantly, participants knew that the AI would help but not how it functions. 

The collaborative search happened by the human first adjusting their dial (including leaving it alone) and then clicking the evaluation button. This initiated the AI helper that took the new participant dial setting and compared two combinations, one using the previous setting of its dial and the other using a new one. The AI was more likely to accept worse values, as well as search distant locations, early on. As the interaction progressed, however, it decreased the likelihood of accepting worse outcomes and its search distance. After evaluating the combinations and selecting one, the new setting was pushed to the feedback window. Control of the interface was then returned to the study participant who had the option of continuing the search or finalizing the choice.

Even though the task is challenging and relatively time-consuming for humans, it is rather simple for most AI. Our research questions are about human behavior when working with an AI helper, not developing better helpers, so we chose to implement a simple agent. The helper uses a stochastic model to determine the likelihood of accepting a worse setting combination and whether to make a big or small leap across the landscape. This is an adaptation of a simulated annealing algorithm. This simple AI ensured richer interactions with the participants and eliminated unwanted variance in the interactions, thus reducing the needed sample size. Lastly, because it could do the task almost instantaneously, we implemented a brief feedback delay to give the sense that the helper was ``working.''

\subsection{Experimental Manipulations}
The two-by-two design crosses gain-loss framing with anchoring. Participants in the gain frame attempt to gather as many points as possible, which go towards their bonus payment, thus their feedback is positive. In the loss frame, participants attempt to reduce the number of points lost, thus the feedback is negative. The left panel of figure \ref{fig:screenshot} depicts a gain frame. The task is to maximize the value in both gain and loss framing. Recall that the landscapes are randomly shifted away from the $[0, 32]$ interval on which they were drawn. Values cover a 33-unit interval in the gain frame on $[0, 100]$, while loss frame values are on $[-100, 0]$. Participants in the anchoring treatment received a message immediately above the dials that informed them of the best possible dial setting's value, i.e., as good as they could do on a given task. Participants without an anchor had to discover what constituted a ``good'' outcome for each task. 

Summarizing the present study: each participant completed four incentivized complex choice tasks, two alone and two with an AI helper, in which they tuned dials to find optimal settings. We did not inform participants of the underlying choice landscapes, which were procedurally drawn and unique to each task. Instead, they were left to uncover insights about the space through experience. Participants did the solo tasks first, but in both the solo and team tasks the order of the one and four peaked landscapes was random. Participants were also randomized into a loss or gain framing treatment conditions. The tuning values of the loss treatment were negative, and participants worked to diminish the number of lost points. Conversely, tuning values of the gain treatment were positive, and participants worked to gain points. The gain-loss framing treatment was crossed with an anchoring treatment in which half of the participants were informed of the best they could find on each task. All participants worked with the same AI helper, but it was stochastic, meaning that even if two participants made the same choice under similar contexts, the helper may have made different choices. Together, these experimental manipulations facilitate studying human-AI teams when the person is operating under a cognitive heuristic or bias. 

\section{Data and Analyses}
We recruited 400 participants via Prolific Academic to complete a study on decision making using dials. We only recruited participants with an approval rate of 100 and who could complete the study on a desktop computer. In the months leading up to data collection, there was an influx of new workers on the platform that significantly skewed the composition of the worker population. Thus, we also restricted participation to workers who joined the platform prior to the influx.\footnote{https://www.prolific.co/blog/we-recently-went-viral-on-tiktok-heres-what-we-learned (We also added this blog to the Wayback Machine at web.archive.org).} Funding and IRB constraints restricted us to recruiting English-speaking American citizens. According to Prolific Academic, this left just over 20,000 eligible workers, roughly $\frac{1}{6}$ of the research pool at recruitment time. The study took about 15 minutes to complete and paid \$2.00 plus a bonus of up to \$2.00 based on effort. The resulting payments yielded an average pay of \$18.40 per hour. 

Of the 400 participants who completed the experiment, 172 participants identified as male, 218 as female, and eight as other. The average age of a participant was 32 years. 203 participants indicated they were college graduates with a four-year degree or higher. We dropped two observations from the data set because the participants failed to complete the study. One participant was a considerable outlier: evaluating 617 settings for one of their challenges, more than the complete set of combinations. Moreover, this is three times the effort of the next most ambitious participant. Since this behavior would likely have an outsized impact on any analysis, we decided to remove this participant from the data set, leaving 397 observations.

We report participants' solo efforts in a companion paper \cite{gurney2022Complex}. We found that when working alone, the four-peak landscapes were more challenging than the one-peak, that both anchoring and framing yielded significant main effects on participants' choice quality, and that doing the one-peak task first was correlated with better outcomes on the four-peak task. Participants in the loss frame submitted more settings, all else equal, and spent more time exploiting versus exploring. Having an explicit anchor, on the other hand, was correlated with submitting fewer dial settings and did not significantly alter participants' search strategies. 

\subsection{Complex Choice with the AI Helper}
\begin{figure}[ht]
    \includegraphics[width=0.475\textwidth]{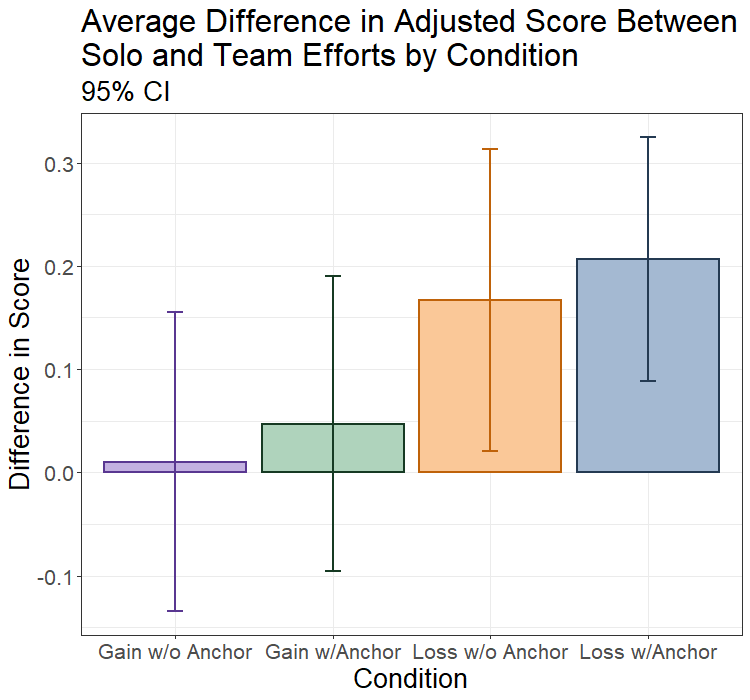}
    \caption{Adjusted score accounts for variance in the landscapes and the increase in average elevation created by adding more peaks. The difference in score is simply the total score achieved during the solo effort minus the total score achieved while working with the AI helper.}
  \label{fig:scoreDiff}
\end{figure}

Each landscape is unique and, because the space of the landscape is fixed, adding more peaks increases the average elevation. Thus, comparing raw performance on the landscapes does not fully depict the impact of working with the AI helper. We account for this by dividing each landscape score by the landscape's average elevation to create an adjusted score. On average, participants' total adjusted score for the two choice tasks was worse when they worked with the AI helper $(M = 3.613, \, SD = 0.717)$ than on their own $(M = 3.718, \, SD = 0.820)$. This difference in total adjusted score $(0.105,\,95\%\, CI\, [0.036, 0.175])$ was statistically significant $(t(396) = 2.985,\, p = 0.003)$ per a paired sample t-test. As illustrated by figure \ref{fig:scoreDiff}, the difference was primarily driven by participants in the loss framing. Whether there was an anchor present $(t(90) = 3.464,\, p < 0.001)$ or not $(t(101) = 2.270,\, p = 0.025)$, the difference in adjusted scores for participants in the loss framing was significant. 

A factorial ANOVA predicting the difference between solo and team efforts in the total adjusted score using the experimental treatments as interacting independent variables revealed a significant main effect for framing but not anchoring or the interaction. Thus, a two-sample t-test is sufficient for comparing the effect of framing on performance. Participants in the loss framing condition achieved significantly better $(t(394.420) = 2.230,\, p = 0.026)$ scores than their counterparts $(M_{difference} = 0.156,\,95\%\, CI\, [0.185, 0.294])$. 

Participants did worse on the four- $(M = 1.522,\, SD = 0.259)$ versus one-peak $(M = 2.091,\, SD = 0.554)$ landscape. This difference $(0.569,\,95\%\, CI\, [0.521, 0.616])$ was significant $(t(396) = 23.431,\, p < 0.001)$. A factorial ANOVA predicting the score difference between the two landscape types using the experimental treatments as interacting independent variables revealed no main effects for anchoring $(F(1, 393) = 2.792,\, p = 0.100)$ and framing $(F(1, 393) = 0.128,\, p = 0.721)$, nor an interaction effect $(F(1, 393) = 0.342,\, p = 0.559)$. 

Relative to their solo efforts $(M = 2.175,\, SD = 0.631)$, participants did worse on the one-peak landscape when working with the AI $(M = 2.091,\, SD = 0.554)$, a difference $(0.084,\,95\%\, CI\, [0.025, 0.144])$ which was statistically significant $(t(396) = 2.773,\, p = 0.006)$. This did not hold for the four-peak landscapes $(t(396) = 1.342,\, p = 0.181)$. A factorial ANOVA predicting the difference between solo and team efforts in the adjusted score for the one-peak landscape using the experimental treatments as interacting, independent variables revealed a significant main effect for the framing treatment condition but not for the anchoring or the interaction. Thus, a two-sample t-test is sufficient for comparing the effect of framing on performance on the one-peak landscapes. Participants in the loss framing condition achieved significantly better $(t(393) = 2.010,\, p = 0.045)$ scores than their counterparts $(M_{difference} = 0.122,\,95\%\, CI\, [0.003, 0.242])$. 

Search duration (number of submitted settings) and the explore/exploit trade-off (fraction of submitted settings that explore new regions, i.e. search strategy) serve as the main behavioral metrics. On average, participants searched less, i.e. submitted fewer dial settings across the two tasks, when they worked with the AI helper $(M = 30.368,\, SD = 27.786)$ than on their own $(M = 43.202,\, SD = 41.521)$. This difference in search duration $(12.834,\,95\%\, CI\, [9.512, 16.155])$, per a paired-sample t-test, was statistically significant $(t(396) = 7.596,\, p < 0.001)$. The fraction of those submissions that were explores while working with the AI helper $(M = 0.612,\, SD = 0.243)$ was greater than when they were working on their own $(M = 0.447,\, SD = 0.232)$. This difference in the explore/exploit trade-off $(-0.165,\, 95\%\, CI\, [-0.186, -0.144])$ was also statistically significant $(t(396) = -15.3456,\, p < 0.001)$. The absolute number of submissions that were explores while working with the AI helper $(M = 13.584,\, SD = 6.244)$, however, was not statistically different from the solo effort $(M = 13.411,\, SD = 8.617;\, t(396) = -0.442,\, p = 0.659)$. This suggests that participants invested less effort in fine-tuning their submissions, perhaps because they anticipated that the AI would do so for them. 

\subsubsection{Search Duration with Anchoring and Framing Effects}
Comparing the difference in search duration for participants' solo and team efforts under the different treatment conditions provides insight into how anchoring and framing effects impact human-AI teams. A factorial ANOVA predicting the difference in search duration (solo effort minus team effort duration) using the experimental treatments as interacting, independent variables revealed significant main effects for both treatment conditions but not the interaction, thus we removed the interaction from the model. The resulting model suggests strong main effects for both anchoring $(F(1, 394) = 14.971,\, p < 0.001)$ and framing $(F(1, 394) = 16.751,\, p < 0.001)$. A post hoc Tukey test showed that the anchoring $(-12.611,\, 95\%\, CI\, [-21.047, -4.174],\, p_{adj} < 0.001)$ and framing $(13.331,\, 95\%\, CI\, [4.892, 21.769],\, p_{adj} < 0.001)$ effects were significant at the $p < 0.01$ level, although search duration moved the opposite way. 

\begin{figure}[ht]
    \includegraphics[width=0.475\textwidth]{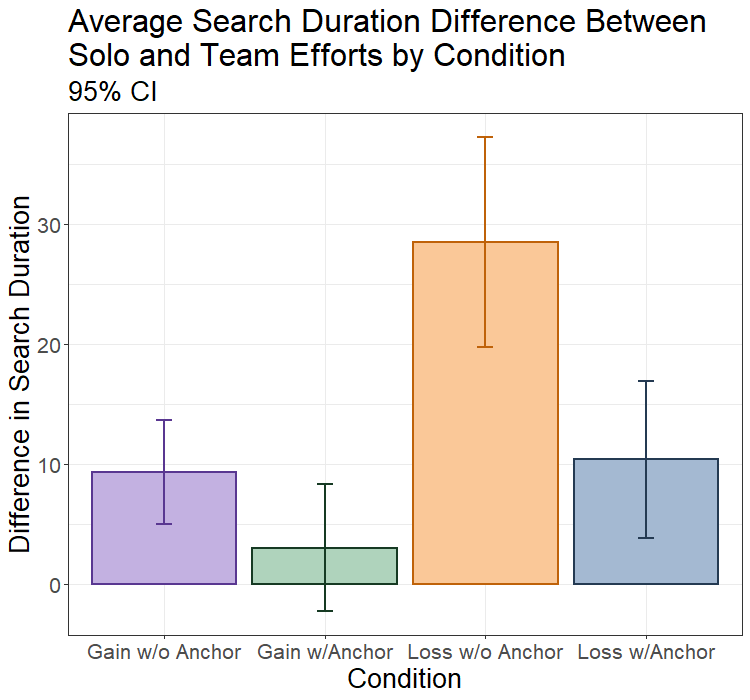}
    \caption{Search duration difference is the total number of submitted settings during a participant's solo effort minus the total number of submitted settings while working with the AI helper. }
  \label{fig:duration}
\end{figure}

Building on these insights, using an independent sample t-test, loss frame participants adjusted their effort downwards  $(M = -19.964,\, SD = 39.838)$ significantly more than gain frame participants $(M = -6.088,\, SD = 24.808;\, t(318.21) = -4.319,\, p < 0.001)$ when they started working with the AI. This is the inverse of the anchoring effect: participants with explicit anchors adjusted their effort downwards $(M = 6.449,\, SD =  29.426)$ significantly less than those without $(M = 19.060,\, SD = 36.341;\, t(382.16) = 3.804,\, p < 0.001)$. As illustrated in figure \ref{fig:duration}, the anchoring effect seems to have tempered the framing effect.

On average, participants searched less in each successive task. However, the only significant drop was between their second solo effort and the first team effort $(M_{difference} = -5.108;\, t(396) = 5.003,\, p < 0.001)$. The difference between the first and second solo $(M_{difference} = -1.690;\, t(396) = 1.654,\, p = 0.099)$ as well as the first and second team $(M_{difference} = -0.927;\, t(396) = 1.300,\, p = 0.194)$ efforts were not significant. This points to participants expecting that the AI would reduce the effort they needed to invest in the task to perform well. For gain-frame participants, this appears to be true: as they exerted less effort while working with the AI $(M_{difference} = -6.088;\, t(203) = -3.505,\, p < 0.001)$, but averaged the same adjusted scores $(M_{difference} = -0.029;\, t(203) = -0.574,\, p = 0.567)$. Loss frame participants, however, appear to have over-adjusted their effort $(M_{difference} = -19.964;\, t(192) = -6,962,\, p < 0.001)$ such that their adjusted scores suffered significantly $(M_{difference} = -0.186;\, t(192) = -3.877,\, p < 0.001)$. For the anchoring treatment, whether a participant saw an anchor $(M_{difference} = -6.449;\, t(195) = -3.068,\, p = 0.002)$ or not $(M_{difference} = -19.060;\, t(200) = -7.436,\, p < 0.001)$, the difference in their effort was significantly less when they worked with the AI helper. This was not correlated with a significantly lower adjusted score during the AI tasks for the no-anchor participants $(M_{difference} = -0.090;\, t(200) = -1.730,\, p = 0.090)$. It was, however, correlated with a significantly lower adjusted score for the anchor participants $(M_{difference} = -0.121;\, t(195) = -2.539,\, p = 0.012)$.

\subsubsection{Search Strategy with Anchoring and Framing Effects}
As noted, the change in search duration echoed through to the strategy participants used when they began working with the AI helper. This shorter search duration meant a higher fraction of submissions that were explorations. A factorial ANOVA predicting the difference in search strategy (solo effort minus team effort strategy) using the experimental treatments as interacting, independent variables revealed no main effects for anchoring $(F(1, 393) = 1.934,\, p = 0.165)$ and framing $(F(1, 393) = 0.684,\, p = 0.409)$, nor an interaction effect $(F(1, 393) = 1.150,\, p = 0.284)$. Again, we interpret this as suggesting that the participants anticipated that the AI would reduce the effort they needed to expend in tuning the dials.

\section{Discussion}
Aversion to algorithmic choice aids is a well-documented phenomenon \cite{dietvorst2015algorithm, burton2020systematic}. The proliferation of both passive and adaptive algorithms in every corner of life, however, is leaving people increasingly accepting of them \cite{logg2019algorithm}. Historically, the experimental methods used to study human-algorithm teams relied on reducing the complexity of choices or studying them \textit{in situ}. Although both methods have merit, they also have major shortcomings, such as not generalizing to complex choices or to choices made in different contexts. The method we developed here maintains choice complexity and is sufficiently abstract to facilitate the generalization of findings. 

Study participants generally made worse choices when they worked with the AI helper than when they worked alone. This result was primarily driven by participants in a loss frame who dramatically decreased their effort once they started working with the AI helper. However, participants in the gain frame performed about the same on their own as with the AI. The decrease in participants' effort was primarily seen in fine-tuning a choice by exploiting local information rather than exploring new options. The only participants that did not significantly reduce their effort when they started working with the AI helper were those in the gain frame \textit{and} with an explicit anchor, i.e. they knew the value of the best possible choice. These results suggest that people were possibly over-reliant on the AI helper, assumed that it was better at the task than it actually was, or, relatedly, their lack of knowledge about how it functioned hindered their ability to team with it. These possibilities point to interesting and important topics for future research. Additionally, they suggest that the well-documented phenomenon of algorithm aversion may not be a stable aspect of human cognition. 

\section{Conclusion}
AI helpers are making their way into every aspect of life---and people appear to be more willing than ever to allow them to help. Historically, the complex choices for which AI can be most helpful have been reduced to simpler analogs for experiments or studied \textit{in situ}. Although simple choice experiments can easily isolate the effects of heuristics and biases, they ignore the fact that many choices are characterized by astounding complexity. \textit{In situ} experimentation overcomes this, but it may produce results that do not generalize. As demonstrated, the dial tuning task maintains the ability to study well-documented cognitive artifacts during complex choices plus, we argue, it provides general results (and data) that AI systems can use. Specifically, the data generated by this experiment are easy to translate into layered image representations that lend themselves to deep learning models. With sufficient data points, such models could learn to discriminate between biased and unbiased behavior.

\bibliography{bib} 

\section{Acknowledgments}
\bigskip
\noindent Part of the effort behind this work was sponsored by the Defense Advanced Research Projects Agency (DARPA) under contract number W911NF2010011. The content of the information does not necessarily reflect the position or the policy of the U.S. Government or the Defense Advanced Research Projects Agency, and no official endorsements should be inferred. The work was or is also sponsored by the U.S. Army Research Laboratory (ARL) under contract number W911NF-14-D-0005. Statements and opinions expressed and content included do not necessarily reflect the position or the policy of the Government, and no official endorsement should be inferred.

\end{document}